\documentclass[conference,vietnamese]{IEEEtran}
\IEEEoverridecommandlockouts
\usepackage{cite}
\usepackage{hyperref}
\usepackage{amsmath,amssymb,amsfonts}
\usepackage{algorithmic}
\usepackage{graphicx}
\usepackage{textcomp}
\usepackage{xcolor}
\usepackage{float}
\usepackage{array}
\usepackage[utf8]{inputenc}
\usepackage[T5]{fontenc}
\usepackage{multirow}

\usepackage{authblk}

\usepackage{xcolor}
\usepackage{pgfplots}
\usepackage{tikz}
\definecolor{bblue}{HTML}{4F81BD}
\definecolor{rred}{HTML}{C0504D}
\definecolor{ggreen}{HTML}{9BBB59}
\definecolor{ppurple}{HTML}{9F4C7C}

\def\BibTeX{{\rm B\kern-.05em{\sc i\kern-.025em b}\kern-.08em
    T\kern-.1667em\lower.7ex\hbox{E}\kern-.125emX}}
\begin{document}

\title{Job Prediction: From Deep Neural Network Models to Applications}

%\author{\textbf{Tin Van Huynh}, %\textbf{Kiet Van Nguyen}, \textbf{Ngan Luu-Thuy Nguyen}, \textbf{Anh Gia-Tuan Nguyen}\\ University of Information Technology, Ho Chi Minh City, Vietnam\\ Vietnam National University, Ho Chi Minh City, Vietnam\\16521827@gm.uit.edu.vn, \{kietnv, ngannlt, anhngt\}@uit.edu.vn}

\author[1,2,*]{Tin Van Huynh}
\author[1,2,†]{Kiet Van Nguyen}
\author[1,2,†]{Ngan Luu-Thuy Nguyen}
\author[1,2,†]{Anh Gia-Tuan Nguyen}
\affil[1]{University of Information Technology, Ho Chi Minh City, Vietnam}
\affil[2]{Vietnam National University, Ho Chi Minh City, Vietnam}
\affil[ ]{Email: *16521827@gm.uit.edu.vn, †\{kietnv, ngannlt, anhngt\}@uit.edu.vn}

\maketitle

\begin{abstract}
Determining the job is suitable for a student or a person looking for work based on their job descriptions such as knowledge and skills that are difficult, as well as how employers must find ways to choose the candidates that match the job they require. In this paper, we focus on studying the job prediction using different deep neural network models including TextCNN, Bi-GRU-LSTM-CNN, and Bi-GRU-CNN with various pre-trained word embeddings on the IT job dataset. In addition, we proposed a simple and effective ensemble model combining different deep neural network models. Our experimental results illustrated that our proposed ensemble model achieved the highest result with an F1-score of 72.71\%. Moreover, we analyze these experimental results to have insights about this problem to find better solutions in the future.
\end{abstract}

\begin{IEEEkeywords}
    Job Prediction, TextCNN, Bi-GRU-LSTM-CNN, Bi-GRU-CNN, Deep Neural Network
\end{IEEEkeywords}

\section{\textbf{Introduction}} \label{introduction}
In recent years, the strong development of Information Technology (IT) has led to a variety of job positions as well as the requirements of each type of IT job. With the diversity, students or job seekers find the job suitable for their knowledge and skills accumulated at the school or in the process of working are challenging. Also, the recruitment company must filter the profiles of the candidates manually to choose the people suitable for the position they are recruiting, causing a lot of time while the number of applications could be increased to hundreds or thousands. Therefore, we would like to study the task of IT job prediction to help them effectively address the aforementioned issues.

Job prediction is a classification task using several techniques in machine learning and natural language processing trying to predict a job based on job descriptions including job requirements, knowledge, skills, interests, etc. In this paper, we focus on studying on job descriptions collected specifically from the online finding-job sites. In particular, we are interested in IT job descriptions. The task is presented as follows.
\begin{itemize}
\item \textbf{Input}: Given an IT job description collected from the online finding-job sites.
\item \textbf{Output}: A predicted job title for this description.
\end{itemize}
Several examples are shown in Table \ref{tab:examples}. \\
\begin{table}[!ht]
\caption{Several examples for input and output of IT Job Prediction}
\label{tab:examples}
\begin{tabular}{|p{6cm}|l|}
\hline
\multicolumn{1}{|c|}{\textbf{Input}} & \multicolumn{1}{|c|}{\textbf{Ouput}} 
\\ \hline
This is where you come in. We need your help to generate data-driven solutions to our business problems. We need someone who understands mathematical analysis and knows how to use technology to implement it. You’re someone who’s always itching to solve the problem. You’re never satisfied with a feathery answer. You want the facts. You always back up your position with data. & \begin{tabular}[c]{@{}l@{}}Data \\ scientist\end{tabular}          \\ \hline
You will get an opportunity to work with a very talented team to modernize our platform and/or possibly work directly with clients on our services engagements. Requirements: Bachelor degree in Computer Science or equivalent development experience 3+ years of experience with Java/J2EE. Experienced in SQL + Hibernate. Good working knowledge of HTML5, JavaScript, CSS3, XML, Web services, Struts. Experience with Angular 2 or 4 preferred Experience with application servers like Tomcat and WebLogic preferred Good software design understanding Ability to work in team in diverse/ multiple stakeholder environment This position has the potential to interact with clients and stakeholders, therefore must have excellent communication and interpersonal skills Great attention to detail. & \begin{tabular}[c]{@{}l@{}}Full \\ Stack \\ Developer\end{tabular} \\ \hline
\end{tabular}
\end{table}

As in this paper, our four key contributions are summarized as follows.
\begin{itemize}
\item Firstly, we conducted various experiments on the IT job dataset. In particular, we compared different experimental results on deep learning models such as TextCNN, Bi-GRU-LSTM-CNN, Bi-GRU-CNN with various pre-trained word embeddings. The Bi-GRU-CNN model achieved the highest results among the three methods.
\item Secondly, we proposed a simple and effective ensemble model combining different deep neural networks for this classification problem. The performance of this model is higher than single models.
\item Thirdly, we conducted a detailed analysis of the experimental results. In addition, we propose directions for future work based on unresolved challenges.
\item Lastly, we build an application that supports users to enter skills, knowledge and interest, and the results returned by this application is a suitable job suggestion. It is really helpful for anyone who want to find a job.
\end{itemize}

The structure of the paper is organized as follows. Related documents and studies are presented in Section \ref{Related Work}. The IT job dataset is described in Section \ref{dataset}. Section \ref{wsmethod} describes the methods we implement. The experimental results and analysis are presented in Section \ref{experiment}. We also introduce the job-prediction application shown in Section \ref{app}. Conclusion and future work are deduced in Section \ref{conclusion}.
\section{\textbf{Related Work}}
\label{Related Work}

TextCNN demonstrates its effectiveness in machine learning in general and in natural language processing in particular. Firstly, we want to mention the CNN model for text classification problem \cite{Yoon} that has been conducted experiments and evaluated on a variety of datasets such as MR \cite{Pang}, SST-1, SST-2 \cite{Socher}, Subj \cite{Pang1}, TREC \cite{Li}, CR \cite{Hu}, MPQA \cite{Wiebe}, VLSP-2018 \cite{van2018deep} and UIT-VSMEC \cite{ho2019emotion} and obtained quite good performances, or this use in for Hate Speech Detection \cite{Sikdar,Tin} gives good results. This model also proves effective with problems similar to image classification \cite{Krizhevsky}. Besides, in order to increase the model's predictive results, words are represented in a vector space to express well the semantic relationship between words together using pre-trained word embeddings \cite{P.Wang} such as Glove \cite{Pennington} and FastText \cite{Bojanowski}.

In addition, we also would like to study combination models. The classification method in which a combination of different models is an important technique for text classification. Because this method of classification shows a good optimization of the predicted results, better results other than simple models. For instance, there are  several studies such as the Bi-RNN model\cite{Schuster}, the Bidirectional-LSTM model\cite{Zhou,hang2019,phu2018}, the Bidirectional-GRU model\cite{Lu}, the Bi-LSTM-CRF model\cite{Z.Huang}, and the Bi-LSTM-CNN model\cite{Li1}.

A machine learning model with a classification task that can have multiple outputs is challenging. There are the classifier ensemble methods combining multiple outputs of multiple models to increase the performance of the prediction. There are many basic \cite{Remya,Hagen} to advanced ensemble methods \cite{Wolpert, Oza}. However, in this paper, we use the majority voting method. This method is quite simple and recently proved effective in text classification problems \cite{Onan, Remya, Usman}.
\section{\textbf{Dataset}}
\label{dataset}

In this paper, we use the dataset for IT job prediction proposed by Papachristou \cite{Papachristou}. This dataset consists of 10,000 distinct job descriptions collected from the online finding-job sites, annotated with 25 different types of IT-related job. Those categories are mostly related to roles that we typically find in the data-driven economy of today. Table \ref{tab:2} presents the distribution of IT job labels in the Papachristou's dataset. The rate of each label is from 3.20\% to 5.11\%. The dataset is quite balanced. However, there is a lot of overlapping information between the descriptions of these jobs. This is also a challenging dataset for us to find the best model.

\begin{table}[H]
\centering
\caption{Statistics of Papachristou’s Dataset}
\label{tab:2}
\begin{tabular}{|c|l|r|r|}
\hline
\textbf{No.} & \textbf{IT Job Title}                        & \textbf{\#Samples} & \textbf{Percentage} (\%) \\ \hline
1   & Data Scientist                & 400               & 4.00               \\ \hline
2   & Data Analyst                  & 397               & 3.97            \\ \hline
3   & Data Architect                & 399               & 3.99            \\ \hline
4   & Data Engineer                 & 385               & 3.85            \\ \hline
5   & Statistics                    & 390               & 3.90             \\ \hline
6   & Database Administrator        & 400               & 4.00               \\ \hline
7   & Business Analyst              & 396               & 3.96            \\ \hline
8   & Data and Analytics Manager    & 399               & 3.99            \\ \hline
9   & Machine Learning              & 392               & 3.92            \\ \hline
10  & Artificial Intelligence       & 382               & 3.82            \\ \hline
11  & Deep Learning                 & 381               & 3.81            \\ \hline
12  & Business Intelligence Analyst & 372               & 3.72            \\ \hline
13  & Data Visualization Expert     & 377               & 3.77            \\ \hline
14  & Data Quality Manager          & 394               & 3.94            \\ \hline
15  & Big Data Engineer             & 320               & 3.2             \\ \hline
16  & Data Warehousing              & 385               & 3.85            \\ \hline
17  & Technology Integration        & 399               & 3.99            \\ \hline
18  & IT Consultant                 & 398               & 3.98            \\ \hline
19  & IT Systems Administrator      & 399               & 3.99            \\ \hline
20  & Cloud Architect               & 540               & 5.4             \\ \hline
21  & Technical Operations          & 394               & 3.94            \\ \hline
22  & Cloud Services Developer      & 395               & 3.95            \\ \hline
23  & Full Stack Developer          & 400               & 4.00               \\ \hline
24  & Information Security Analyst  & 395               & 3.95            \\ \hline
25  & Network Architect             & 511               & 5.11            \\ \hline
\multicolumn{2}{|c|}{Total}         & 10,000             & 100             \\ \hline
\end{tabular}
\end{table}
% Please add the following required packages to your document preamble:
% \usepackage{booktabs}
% \usepackage{longtable}
% Note: It may be necessary to compile the document several times to get a multi-page table to line up properly
% Please add the following required packages to your document preamble:
% \usepackage{longtable}
% Note: It may be necessary to compile the document several times to get a multi-page table to line up properly

\section{\textbf{Methodology}}
\label{wsmethod}
\begin{figure*}[]
\centering
  \includegraphics[scale=0.5]{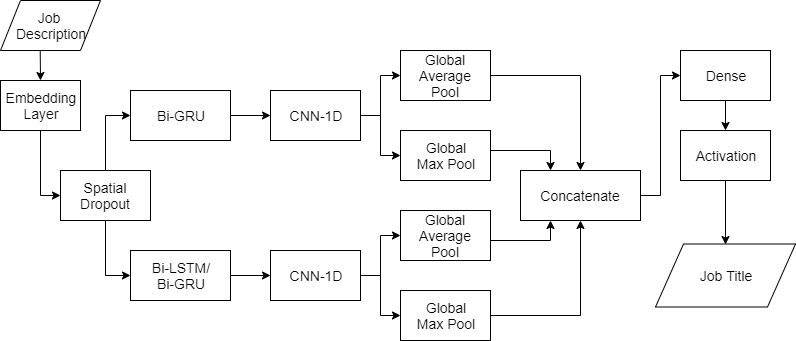}
  \caption{The Bi-GRU-CNN or Bi-GRU-LSTM-CNN architecture for the job classification}
  \label{fig:model1}
\end{figure*}
In this study, we implemented four deep neural network models for the IT job classification, a single model TextCNN, two combination models (Bi-GRU-CNN and Bi-GRU-LSTM-CNN), and our proposed ensemble model. In addition, we implement two pre-trained word embeddings into these models.
\subsection{TextCNN}
Firstly, we motivated from TextCNN which is proposed by \cite{Yoon} to classify job descriptions. TextCNN is a deep learning algorithm that achieves the best results in many studies of Natural Language Processing which includes both emotion recognition, sentiment analysis, and question classification.

The TextCNN model consists of three primary parts such as convolution layer, pooling layer, and Fully Connected layer. In Convolution layer - the Kernel, we used 3 types of filters of different sizes with a total 512 filters to extract the high-level features and obtain convolved feature maps. These then go through the Pooling layer which is responsible for reducing the spatial size of the convolved feature and decreasing the computational power required to process the data through dimensionality reduction. The convolutional layer and the Pooling Layer together form the $i^{th}$ layer of a Convolutional Neural Network. Moving on, the final output will be flattened and fed to a regular neural network in the Fully Connected Layer for classification purposes using the softmax classification technique.
\subsection{Bi-GRU-CNN model}
Also, we are inspired by the Bi-GRU-CNN model, which is used in the salary prediction problem \cite{Z.Wang}, 2019 used to predict wages, based on job descriptions such as job content, job requirements, working time, job position, and type of job. The architecture of this Bi-GRU-CNN model is presented in Figure \ref{fig:model1}. The main components of the model are presented below.

\begin{itemize}
\item Word representation layer: The input is a matrix with 1,200x300 dimensions. In particular, each job description has only 1,200 words and each word is represented by a 300-dimensional word embedding. The pre-training word-level vector already is a kind of word representation for deep neural network models since Glove \cite{Pennington}. In our experiments, we choose FastText \cite{Bojanowski} as our pre-training model.
\item CNN-1D layer: In this architecture, we use a 1D spatial drop out with the dropout rate of 0.2. It can avoid over-fitting and to get better generalizations for these models.
\item Bidirectional GRU: As in Figure \ref{fig:model1}, we build two Bidirectional GRU in parallel, and each Gated Recurrent Units (GRU) is without output gate. \cite{K.Cho} proposed by Cho et al. in 2014. Also, GRUs have two other gates such as an update gate and a reset gate. In particular, an update gate is responsible for combining new input with the previous one and the update gate is responsible for how much the previous memory is required to be saved.
\end{itemize}

\subsection{Bi-GRU-LSTM-CNN model}

In addition, We use the Bi-GRU-LSTM-CNN model again in this classification problem. This model proposed by Huynh et al. \cite{Tin} to solve the Hate-speech detection problem  in the VLSP Share Task 2019, achieving the fifth rank on the scoreboard on the public test \cite{Son}. Figure \ref{fig:model1}  presents our proposed network structure. The basic architecture in this model is CNN-1D. In addition, we also studied two other deep neural models, Long Short-Term Memory (LSTM) and Gated Recurrent Unit (GRU). Basically, the Bi-GRU-LSTM-CNN model has the same structure as the Bi-GRU-CNN model, instead of two Bi-GRU in parallel, this model uses a combination of Bi-GRU and Bi-LSTM in parallel. Therefore, we only explain the Bi-LSTM component instead of the whole model as follows.

\begin{itemize}

\item Bidirectional LSTM: The model uses two parallel blocks of Bidirectional Long Short Term Memory (Bi-LSTM) where the term Bidirectional is that the input sequence is given to the LSTM in two different ways. LSTM is a variation of a recurrent neural network
that has an input gate, an output gate, a forget gate and a cell. In our experiment, we used two parallel bidirectional LSTM blocks having 112 units for each. We used sigmoid and tanh for recurrent activations and hidden units respectively
\end{itemize}

\subsection{A Simple Ensemble Model using Majority Voting}
\begin{figure}
\centering
  \includegraphics[scale=0.4]{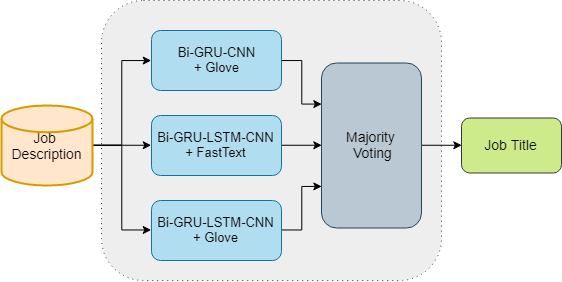}
  \caption{Our proposed ensemble for the job prediction}
  \label{fig:model2}
\end{figure}

In this paper, we use the Majority Voting method to increase the predictive efficiency of the classification model. The structure of the model is shown in Figure \ref{fig:model2}. In this technique, the output of n different models will be used. The final classification result for the problem is the combination of the above outputs by voting. vote for each predicted point. The output tag will be the one that gets the most votes.

In the paper, we use three different m job classification models and are the three models with the greatest accuracy of the models we tested, we predict the y label for each job description through majority (plurality) voting of each C$_{i}$ classification model.
 
$\mathrm{\hat{y}=mode\ \left\{C_1\left(x\right),C_2\left(x\right),\ldots,C_m\left(x\right)\ \right\}}$

For example, we have a sample result as follows:
\begin{itemize}
\item Model 1 -> Label 1
\item Model 2 -> Label 0
\item Model 3 -> Label 1
\end{itemize}
 
 $\mathrm{\hat{y}=mode\ \left\{1,0,1\ \right\}\ =\ 1}$

Through majority voting, we will classify the sample into class 0. In the case of unable to find the result of the label after voting because maybe each model for each label is different, we will choose the last label as the label of the model. Figure \ref{fig:model2} gives the best classification results among the three models.
\section{\textbf{Experiments}}
\label{experiment}
% Please add the following required packages to your document preamble:
% \usepackage{longtable}
% Note: It may be necessary to compile the document several times to get a multi-page table to line up properly
\subsection{Pre-processing}
In this study, we implement several simple and effective techniques to pre-process data for the model's input as follows.
\begin{itemize}
\item Converting the job descriptions into the lowercase strings.
\item Deleting special characters such as \#, \@, \& *, \$, etc.
\item Segmenting job descriptions into a set of words
\item Removing the stop word in the descriptions
\item Representing words into vectors with pre-trained word embedding sets.
\end{itemize}

\subsection{Experimental Settings}
In this study, we split randomly the dataset into three different sets including 10\% for the testing set, 90\% for training set. In the training set, we get 20\% for the validation set. To evaluate our models, we use four measures of measurement such as accuracy, precision, recall, and F1-score.

\subsection{Experimental Results}
\begin{table*}[!h]
\label{expresult}
\centering
\caption{Experimental Results of Different Models}
\begin{tabular}{|c|c|c|c|c|}
\hline
\textbf{Models}                            & \textbf{Accuracy} & \textbf{Precision} & \textbf{Recall} & \textbf{F1-score} \\ \hline
\textbf{FNN \cite{Papachristou}}                              & 62.50              & -                  & -               & -                 \\ \hline
\textbf{Sequential \cite{Papachristou}}                       & 65.65             & -                  & -               & -                 \\ \hline
\textbf{Sequential + Embedding \cite{Papachristou}}           & 34.25             & -                  & -               & -                 \\ \hline
\textbf{TextCNN \cite{Papachristou}}                              & 66.00              & -                  & -               & -                 \\ \hline
\textbf{TextCNN + FastText (300)}             & 69.00             & 69.57              & 68.86           & 69.21             \\ \hline
\textbf{TextCNN + Glove (300)}                & 65.30             & 65.42              & 65.61           & 65.42             \\ \hline
\textbf{Bi-GRU-CNN + FastText (300)}      & 70.20             & 70.69              & 69.94           & 70.31             \\ \hline
\textbf{Bi-GRU-CNN + Glove (300)}         & \textbf{72.40}    & \textbf{72.46}     & \textbf{72.30}  & \textbf{72.38}    \\ \hline
\textbf{Bi-GRU-LSTM-CNN + FastText (300)} & 71.20    & 71.89     & 71.07  & 71.48    \\ \hline
\textbf{Bi-GRU-LSTM-CNN + Glove (300)}    & 70.30    & 70.91     & 70.52  & 70.71    \\ \hline
\textbf{Ensemble}                        & \textbf{72.70}    & \textbf{72.83}     & \textbf{72.59}  & \textbf{72.71}    \\ \hline
\end{tabular}
\end{table*}
We conducted a series of experiments and our experimental results are shown in Table III. In particular, we have noticed that the Bi-GRU-CNN with the word embedding Glove has proven its robustness and outperform other models (Bi-GRU-LSTM-CNN and the TextCNN model). In particular, the result of this model achieves an accuracy of 72.40\%, along with this model also gives impressive results in other measurements such as 72.38\% for F1-score, 72.46\% for precision and 72.30\% for recall. The Bi-GRU architecture combine with CNN has been more effective than the Bi-GRU-LSTM architecture when combined with the CNN model.

Although not the best model in the test set, the Bi-GRU-LSTM-CNN model combined with the pre-trained word embeddings such as Glove and FastText also produced quite high results, and was much higher the results from Papachristou's experiments, his highest result for accuracy is 66.00\% with TextCNN model \cite{Papachristou}, compared to the model Bi-GRU-LSTM-CNN with the word embedding FastText for an accuracy of 71.20\% and the model Bi-GRU- LSTM-CNN with the word embedding Glove giving an accuracy of 70.30\%.

Finally, we conduct experiments on our proposed ensemble model. We found that the ensemble method achieve the best results and was much higher than other models, namely 72.70\% for accuracy, 72.71\% for F1-score, 72.83\% for precision and 72.59\% for recall. It can be seen that the ensemble method has stable results in all metrics.

% Please add the following required packages to your document preamble:
% \usepackage{longtable}
% Note: It may be necessary to compile the document several times to get a multi-page table to line up properly
\subsection{Result Analysis}

When we observed the results of the classification results on each label, the highest results were Bi-GRU-CNN + Glove and the ensemble model presented in Table IV. In the Bi-GRU-CNN + Glove model, the class "IT Consultant" achieved  the largest accuracy of 91.43\% and the class "Data Architect" accounted for the lowest accuracy of 44.19\%. As for the ensemble model, the label with the most accurate classification is "Deep Learning" with 94.44\% and the label "Data Architect" with the lowest accuracy is 43.18\%.
% Please add the following required packages to your document preamble:
% \usepackage{multirow}
\begin{table*}[!ht]
\centering
\caption{Experimental results on each label of the Bi-GRU-CNN + Glove model and Ensemble model}
\begin{tabular}{|c|c|c|c|c|c|c|c|}
\hline
\textbf{No.} & \textbf{Job Title}       & \multicolumn{3}{c|}{\textbf{Bi-GRU-CNN+Glove}}         & \multicolumn{3}{c|}{\textbf{Our Proposed Ensemble Model}}                                                             \\ \cline{3-8} 
                             &                                       & \textbf{Precision} & \textbf{Recall} & \textbf{F1-Score} & \textbf{Precision} & \multicolumn{1}{l|}{\textbf{Recall}} & \multicolumn{1}{l|}{\textbf{F1-Score}} \\ \hline
1                            & \textbf{Artificial Intelligence}      & 74.42              & 86.49           & 80.00             & 71.11              & 86.49                                & 78.05                                  \\ \hline
2                            & \textbf{Big Data Engineer}            & 53.33              & 22.22           & 31.37             & 50.00              & 27.78                                & 35.71                                  \\ \hline
3                            & \textbf{Business Analyst}             & 64.15              & 82.93           & 72.34             & 67.39              & 75.61                                & 71.26                                  \\ \hline
4                            & \textbf{Business Intelligence Analys} & 74.29              & 68.42           & 71.23             & 79.41              & 71.05                                & 75.00                                  \\ \hline
5                            & \textbf{Cloud Architect}              & 71.69              & 76.00           & 73.79             & 67.86              & 76.00                                & 71.69                                  \\ \hline
6                            & \textbf{Cloud Services Developer}     & 78.85              & 74.55           & 76.64             & 77.19              & 80.00                                & 78.57                                  \\ \hline
7                            & \textbf{Data Analyst}                 & 51.22              & 51.22           & 51.22             & 50.00              & 51.22                                & 50.60                                  \\ \hline
8                            & \textbf{Data and Analytics Manager}   & 48.65              & 62.07           & 54.55             & 58.62              & 58.62                                & 58.62                                  \\ \hline
9                            & \textbf{Data Architect}               & 44.19              & 55.88           & 49.35             & 43.18              & 55.88                                & 48.72                                  \\ \hline
10                           & \textbf{Data Engineer}                & 78.38              & 85.29           & 81.69             & 77.78              & 82.35                                & 80.00                                  \\ \hline
11                           & \textbf{Data Quality Manager}         & 81.25              & 63.41           & 71.23             & 80.00              & 68.29                                & 73.68                                  \\ \hline
12                           & \textbf{Data Scientist}               & 75.56              & 64.15           & 69.39             & 72.09              & 58.49                                & 64.58                                  \\ \hline
13                           & \textbf{Data Visualization Expert}    & 75.68              & 73.68           & 74.67             & 74.36              & 76.32                                & 75.32                                  \\ \hline
14                           & \textbf{Data Warehousing}             & 63.82              & 66.67           & 65.22             & 58.82              & 66.67                                & 62.50                                  \\ \hline
15                           & \textbf{Database Administrator}       & 76.92              & 83.33           & 80.00             & 82.86              & 80.56                                & 81.69                                  \\ \hline
16                           & \textbf{Deep Learning}                & 90.24              & 80.43           & 85.06             & \textbf{94.44}     & 73.91                                & 82.93                                  \\ \hline
17                           & \textbf{Full Stack Developer}         & 87.50              & 85.37           & 86.42             & 90.24              & 90.24                                & 90.24                                  \\ \hline
18                           & \textbf{Information Security Analyst} & 78.26              & 78.26           & 78.26             & 82.22              & 80.43                                & 81.32                                  \\ \hline
19                           & \textbf{IT Consultant}                & \textbf{91.43}     & \textbf{91.43}  & \textbf{91.43}    & 86.84              & \textbf{94.29}                       & \textbf{90.41}                         \\ \hline
20                           & \textbf{IT Systems Administrato}      & 78.38              & 85.29           & 81.69             & 75.68              & 82.35                                & 78.87                                  \\ \hline
21                           & \textbf{Machine Learning}             & 62.00              & 79.49           & 69.66             & 63.04              & 74.36                                & 68.24                                  \\ \hline
22                           & \textbf{Network Architect}            & 75.00              & 82.50           & 78.57             & 79.07              & 85.00                                & 81.93                                  \\ \hline
23                           & \textbf{Statistics}                   & 73.68              & 77.78           & 75.68             & 74.36              & 80.56                                & 77.33                                  \\ \hline
24                           & \textbf{Technical Operations}         & 75.00              & 65.63           & 70.00             & 78.57              & 68.75                                & 73.33                                  \\ \hline
25                           & \textbf{Technology Integration}       & 87.50              & 65.12           & 74.67             & 85.71              & 69.77                                & 76.92                                  \\ \hline
\end{tabular}
\end{table*}

We also tried to analyze the predictability of the model according to the length of the descriptions. The results are shown in Figure 3. We found that if the description length is longer, the predictability of the model is lower. We selected our four highest models to observe, it can be seen that all models give better F1-score when job descriptions are less than 500 words, when the job description is above 500 words, the F1-score becomes smaller, indicating that the length of the input description affects the predictive results.
\begin{figure}[H]
    \centering
    \label{fig:my_label}
\begin{tikzpicture}
    \begin{axis}[
        width  = 0.5*\textwidth,
        height = 7cm,
        major x tick style = transparent,
        ybar=2*\pgflinewidth,
        bar width=14pt,
        ymajorgrids = true,
        ylabel = {F1-score (\%)},
        symbolic x coords={0-500, >500},
        xtick = data,
        scaled y ticks = false,
        enlarge x limits=0.5,
        ymin=0,
        legend cell align=left,
        legend style={
                at={(1,1.05)},
                anchor=south east,
                column sep=1ex
        }
    ]
        \addplot[style={bblue,fill=bblue,mark=none}]
            coordinates {(0-500, 73.02) (>500,61.65)};

        \addplot[style={rred,fill=rred,mark=none}]
             coordinates {(0-500,71.55) (>500,61.35)};

        \addplot[style={ggreen,fill=ggreen,mark=none}]
             coordinates {(0-500,71.61) (>500,58.12)};

        \addplot[style={ppurple,fill=ppurple,mark=none}]
             coordinates {(0-500,73.69) (>500,61.78)};

        \legend{Bi-GRU-CNN + Glove,Bi-GRU-LSTM-CNN + FastText,Bi-GRU-LSTM-CNN + Glove,Ensemble}
    \end{axis}
\end{tikzpicture}
    \caption{Experimental results from with different question length}
\end{figure}
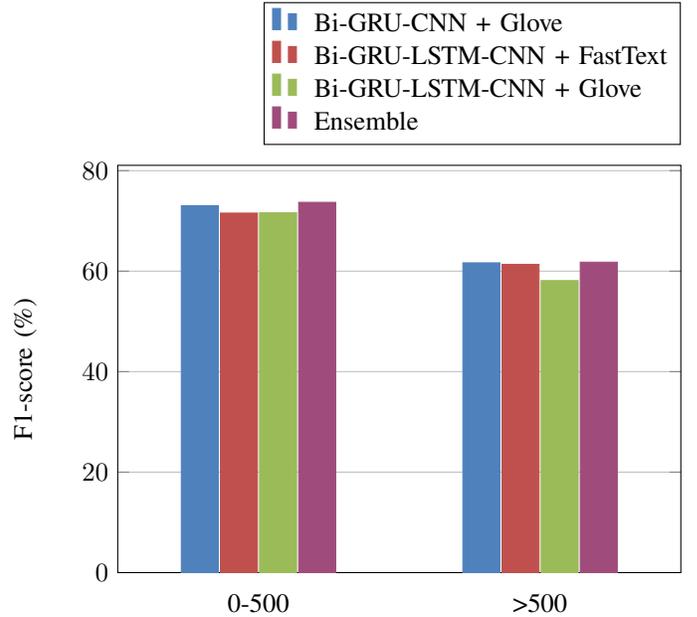

\section{\textbf{Application}}
\label{app}

\begin{figure*}[!ht]
\centering
  \includegraphics[scale=0.7]{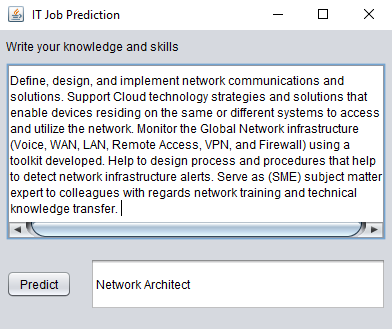}
  \caption{The IT job prediction application demo}
  \label{fig:tool}
\end{figure*}

From our research results, we could implement our highest-performing model to build a few applications such as job prediction applications based on descriptions of user knowledge and skills and the CV filtering application of candidates suitable for job postings. In this paper, we built a simple application as an example for this. In particular, we created a job prediction application which help IT students find a job suitable for their knowledge, skill, interests, etc. Figure \ref{fig:tool} shows our proposed application.

\section{\textbf{Conclusion and future work}}
\label{conclusion}
In this paper, we implemented the TextCNN model and more complex models such as Bi-GRU-LSTM-CNN and Bi-GRU-CNN with various word embeddings to solve the IT job prediction. From experimental results of the deep neural networks and leveraging the power of each model, we also proposed a simple and effective ensemble model and we proved that this method is the best performance for the task of predicting IT jobs. In particular, we achieved the best performance with the F1-score of 72.71\% on the IT-job dataset on our proposed ensemble.

In the future, we would like to improve the quantity as well as the quality of the dataset. In particular, the dataset contains very limited data with 10,000 annotated job descriptions. Furthermore, we aim to experiment with other traditional classifiers with different features and deep learning models with various word representations or combine both methods on this corpus. In order to have a comprehensive view in this research, we will compare traditional machine learning with deep learning on this dataset as the previous work \cite{phu2018}. Besides, we will study the LSTM variants as the study \cite{vu2018} has introduced.

\medskip

\section*{Acknowledgment}
We would like to give our thanks to the NLP@UIT research group and the Citynow-UIT Laboratory of the University of Information Technology - Vietnam National University Ho Chi Minh City for their supports with pragmatic and inspiring advice.

\end{document}